\documentclass[lettersize,journal]{IEEEtran}
\usepackage{amsmath,amsfonts}
\usepackage{algorithmic}
\usepackage{algorithm}
\usepackage{array}
\usepackage[caption=false,font=normalsize,labelfont=sf,textfont=sf]{subfig}
\usepackage{textcomp}
\usepackage{stfloats}
\usepackage{url}
\usepackage{doi}
\usepackage{verbatim}
\usepackage{graphicx}

\usepackage{cite}
\usepackage{multirow}
\usepackage{verbatim}
\usepackage{color}
\usepackage{threeparttable}
\usepackage{diagbox}

\begin{document}

\title{Concealed Object Detection for Passive Millimeter-Wave Security Imaging Based on Task-Aligned Detection Transformer}

\author{Cheng Guo,
        Fei Hu,
        and Yan Hu 
\thanks{Manuscript received XX XX, 2022; revised XX XX, 2022. This work was supported in part by the National Natural Science Foundation of China (NSFC) under Grants 61871438. \emph{(Corresponding author: Fei Hu.)}}
\thanks{C. Guo, F. Hu, and Y. Hu are with the School of Electronic Information and Communications, Huazhong University of Science and Technology, Wuhan 430074, China, and F. Hu are also with the National Key Laboratory of Science and Technology on Multi-Spectral Information Processing, Wuhan 430074, China (e-mail: ch3ngguo@foxmail.com, hufei@hust.edu.cn, dawalish@foxmail.com ).}}


\maketitle

\begin{abstract}
 Passive millimeter-wave (PMMW) is a significant potential technique for human security screening. Several popular object detection networks have been used for PMMW images. However, restricted by the low resolution and high noise of PMMW images, PMMW hidden object detection based on deep learning usually suffers from low accuracy and low classification confidence. To tackle the above problems, this paper proposes a Task-Aligned Detection Transformer network, named PMMW-DETR. In the first stage, a Denoising Coarse-to-Fine Transformer (DCFT) backbone is designed to extract long- and short-range features in the different scales. In the second stage, we propose the Query Selection module to introduce learned spatial features into the network as prior knowledge, which enhances the semantic perception capability of the network. In the third stage, aiming to improve the classification performance, we perform a Task-Aligned Dual-Head block that decouples the classification and regression tasks, and enhances the interaction between the two heads. Based on our self-developed PMMW security screening dataset, experimental results including comparison with State-Of-The-Art (SOTA) methods and ablation study demonstrate that the PMMW-DETR obtains higher accuracy and classification confidence than previous works, and exhibits robustness to the PMMW images of low quality.
\end{abstract}


\section{Introduction}
\label{intro}

Passive millimeter-wave (PMMW) imaging has attracted increasing attention from academia and industry in recent years \cite{salmon2017outdoor,yujiri2003passive,salmon2020indoor}. Similar to sensors in the infrared and visible bands, PMMW sensors receive the self-emitted radiation of objects and reflected environmental radiation. PMMW sensors, on the other hand, do not require an irradiation source and have the ability to detect through smoke, dust, and light rain, making PMMW imaging capable of all-day and quasi-all-weather, so it is widely used in astronomy and remote sensing \cite{casella2022can,cuadrado2022solid}. With the 
improvement in resolution brought by the development of millimeter-wave (MMW) devices, MMW radiometry is gradually applied to the information acquisition of close-range objects \cite{su2021detection,cheng2020concealed,cheng2020passive,owda2022passive}. Among these, the most widely used application is the detection of concealed objects such as firearms, gasoline, and knives in human security screening for harmlessness to human safety and the ability to penetrate textile materials of MMW. 

Since the 1990s, researchers have designed a variety of security systems and conducted many human imaging experiments. So far, the performance of the human body security inspection system has gradually met the application requirements. In recent years, Tian \textit{et al.} \cite{tian2012design} develop a security system based on active ElectroMagnetic (EM) detection methods. Gumbmann \textit{et al.} \cite{Gumbmann2017short} proposed a Stepped-frequency continuous-wave (SFCW) radar system with an optimum sweep design. Cheng \textit{et al.} \cite{cheng2021passive,cheng2020passive} first applied polarization imaging mode to the PMMW security system. And Zhao \textit{et al.} \cite{zhao2022novel} develop a Ka-band 1024-channel PMMW security system named BHU-1024. It can be seen that the security systems have reached a relatively mature stage. However, the current system is still not commercially available, the main reason is that the level of concealed object detection algorithms for PMMW security images lags behind the hardware level. The PMMW radiation signal is weak, with only one hundred thousandths of the infrared signal, so the image contains a lot of noise. Meanwhile, the resolution of PMMW images is much lower than that of infrared and visible images, making concealed object detection a challenging task. Many methods have been proposed for the detection of concealed objects in PMMW security images. Earlier research works were mainly based on statistical learning and machine learning, for example, Martinez \textit{et al.} \cite{martinez2010concealed} used Iterative Steering Kernel Regression (ISKR) method for denoising and the Local Binary Fitting (LBF) method to segment hidden objects and the human body. Lee \textit{et al.} \cite{lee2010automatic} applied Bayesian learning and expectation maximization algorithms to detect and segment concealed objects. Yeom \textit{et al.} \cite{yeom2011real,yeom2012real} further proposed the use of principal component analysis (PCA) to extract object features and measure the similarity between the object and ground truth based on Lee's work. Lopez \textit{et al.} \cite{lopez2018using}  extracted Haar features for images and used classical machine learning algorithms such as random forest and support vector machine for detection.

The requirement for manual design features is a common disadvantage of the above methods, which leads to poor generalization ability and does not adapt as well when the object distribution changes. Deep learning has now been very successful in the field of computer vision, and many networks have achieved impressive performance on available datasets. Inspired by these works, deep learning, especially convolutional neural network (CNN), has been attempted to be applied to PMMW image feature extraction and object detection in recent years, which avoids the need to manually design complex features. For example, Lopez \textit{et al.} \cite{lopez2017deep} performed smoothing preprocessing and applied SCNN networks for object segmentation. Mainstream networks in industrial applications, such as YOLOv3 and SSD, have also been used for concealed object detection in PMMW security images \cite{kowalski2019real,pang2020real}. The common shortcoming of the above works is that they all need to set the prior anchor boxes manually, whose shape and size have a significant impact on the final performance of the network.

The attention mechanism \cite{luong2015effective,bahdanau2017neural} is a bionic approach to human vision that allows the detection network to focus on the object site of interest by computing the mutual correlation between sample elements. With its global perception capability and superior detection performance, Transformer excels in the field of computer vision. Typical computer vision Transformer models include the Visual Transformer (ViT) \cite{kolesnikov2021image}, Swin Transformer \cite{liu2021swin}, and Detection Transformer (DETR) \cite{carion2020end}. ViT pioneered the application of Transformer to computer vision tasks by treating the Patch of an image as a vector. However, the computation required by ViT to apply the attention mechanism directly to the full graph is very large, so Swin Transformer proposed a windowed attention mechanism and a shifted windowed attention mechanism to address this problem, reducing the computation of the detection network while weakening the long-range sensing capability of the network. DETR, on the other hand, pioneers an ensemble prediction-based object detection method that eliminates the necessity for post-processing processes such as the manual design of a priori anchor frames and Non-Maximal Suppression (NMS). However, DETR converges very slowly due to the cross-attention mechanism and the instability brought by the randomness of its bipartite graph matching. The design for optical images makes either Swin or DETR perform poorly on the PMMW dataset, therefore it is necessary to design a Transformer that adapts to the characteristics of PMMW imaging.

Yang \textit{et al.} \cite{yang2022transformer} proposed the PMMW-Transformer for human security screening, an Anchor-Free Transformer network that implements adaptive assignment of positive and negative sample labels. The network proposes a new local-global attention mechanism and Attention Weighting Module inspired by Twins \cite{chu2021twins}, Swin, ATSS \cite{zhang2020bridging}, and other networks. However, it does not well address the problems of difficult classification on low-quality images, low AP at high IoU thresholds, and difficult convergence on small datasets. The reasons are mainly the following: 

First, the PMMW security images have low resolution and little texture information, which is easy to cause a low precision rate and low recall rate. Second, due to the intrinsic conflicts between classification and regression \cite{wu2020rethinking}, the network model has difficulty in learning classification information on such a dataset. Third, the inductive bias of Transformer is much weaker than CNN and RNN, which makes the network harder to train on small-scale datasets \cite{xuhong2018explicit}, Especially for data with high noise like PMMW security images. 

To address the characteristics of PMMW security images in security scenarios, this paper proposes a task-aligned detection Transformer network named PMMW-DETR. The backbone of the Transformer network consists of several hierarchical multi-scale Denoising Coarse-to-Fine Transformer (DCFT) attention blocks so as to extract features, and the extracted feature map enters the neck composed of an encoder and a query selection module to further aggregate features. The output features, after introducing a spatial prior, are divided into two independent decoders to perform classification and regression tasks, the queries of both are shared to ensure that they are predicting the same object in the similar regions, which also called task-aligned. The main contributions of this paper are as follows:

1) For the previous problem of classification-regression conflict, we propose a new task-aligned dual-head block design, in which two decoders share queries while performing different (classification/regression) tasks with different parameters, achieving decoupled classification and regression in two different feature spaces.

2) We propose a query selection method adapted to the security check scenario, i.e., introducing a spatial prior to the detection head. Thus, the semantic perception capability of the network is enhanced. In addition, we perform an interaction method which we called query sharing between two decoders for the selected queries.

3) In order to balance global information perception and computational effort, we propose a coarse-to-fine multi-scale DCFT attention block. The coarse global attention is used to help the network understand human body semantic information to reduce false alarms caused by object-similar backgrounds, such as gaps between limbs and torso, while the fine local attention provides rich classification information. In addition, by redesigning the pooling part of the block, we improve the robustness of the network to measurement noise.

4) PMMW-DETR surpasses the popular SOTA models in accuracy. To train our network, we contribute a new dataset that contains 4543 images with high-quality annotations. To the best of our knowledge, this is the first dataset with the best image quality that represents the level of the most advanced PMMW imaging technology. With this dataset, we compared the classification performance of different models for the first time.

\section{Proposed Method}

\begin{figure*}[htp]
\centering
\includegraphics[width=16cm]{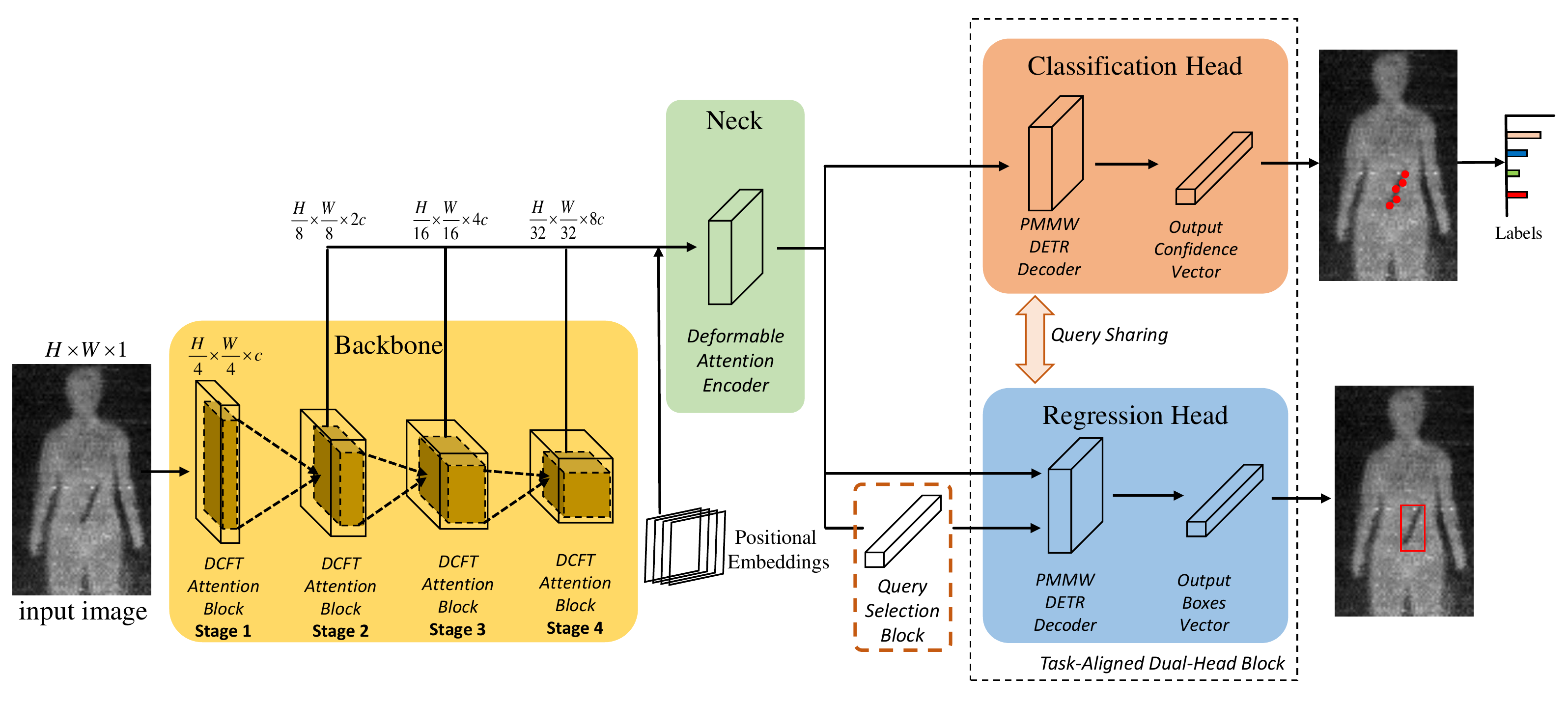}
\caption{The overall structure of the Task-aligned PMMW-DETR.}
\label{overall}
\end{figure*}

\subsection{The Overall Structure}

In order to solve the problem of insufficient information and difficult convergence of the PMMW image dataset, we propose a Transformer-based concealed object detection network named PMMW-DETR. Like the current mainstream detection network, our Transformer network consists of the backbone, neck, and head. As shown in Figure \ref{overall}, the backbone is mainly applied to extract low-level features of images, while the neck functions to further aggregate and extract high-dimensional features, and the head performs classification and localization tasks on the obtained feature maps. For the problems mentioned in Section \ref{intro}, we propose 1) a coarse-to-fine multi-scale DCFT attention block that balances global information and computational efforts, 2) a multi-scale deformable attention module that effectively integrates independent semantic features, and the query selection module that introduces a spatial prior for the detection head, and 3) a task-aligned dual-head block with query sharing that fully aligns the classification and localization tasks, on the backbone, neck, and head, respectively.

\subsection{Denoising Coarse-to-Fine Transformer (DCFT) Attention-Based Backbone}
\label{Sec_backbone}
The attention mechanism has the advantage of long-range perception compared to convolution. The ability to fully extract features of the whole picture makes it well suited for detection tasks with less local information such as PMMW security screening. However, the original attention mechanism does fine-grained operations on the whole globe, resulting in a quadratic computational overhead w.r.t the image size \cite{liu2021swin}, which is unacceptable. Although Swin Transformer proposed the local attention mechanism and the shifted window attention mechanism to reduce the computational cost, the long- and short-range perception capability of the attention mechanism is greatly weakened since the size of the window is far less than that of  the image, which might cause the false alarms due to similar local features between objects and the background.

\begin{figure*}[htp]
\centering
\includegraphics[width=15cm]{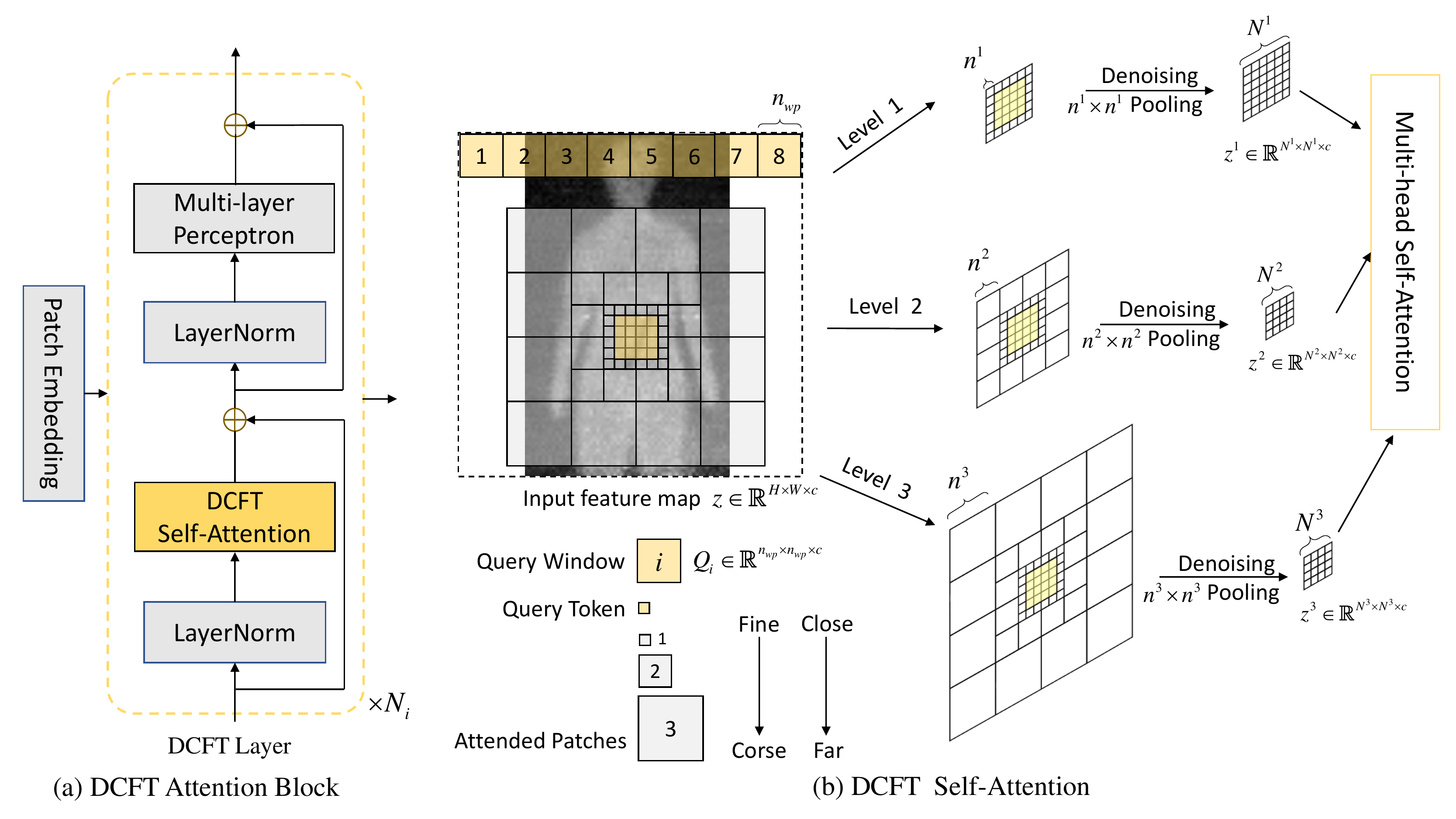}
\caption{An illustration of DCFT Attention Block. (a) A DCFT Attention Block consists of a Patch Embedding layer and $N_{i}$ DCFT layers. (b) An illustration of the DCFT attention at window levels. Each of the finest square cells represents a query token.}
\label{coarse-to-fine}
\end{figure*}

To balance the performance and computational overhead of the attention mechanism, we propose the DCFT attention-based backbone which consists of four DCFT attention blocks, as shown in Figure \ref{overall}. In order to obtain a hierarchical multi-scale backbone structure, we partition each image $I \in \mathbb{R}^{{H} \times {W} } $ into patches of size $4 \times 4$, resulting in $\frac{H}{4} \times \frac{W}{4}$ visual tokens with dimension $4 \times 4 $. Then, as shown in Figure \ref{coarse-to-fine}(a), a patch embedding layer consisting of a convolutional layer with stride and filter size both equal to 4, is used to project the patches into the feature map $z \in \mathbb{R}^{\frac{H}{4} \times  \frac{W}{4} \times c} $. After the projection, similar operations are performed on the feature map in the four stages of DCFT attention blocks. At each stage $i \in\{1,2,3,4\}$, the DCFT attention block consists of $N_{i}$ DCFT layers. In particular, for the patch embedding layer in the $i \in \{2,3,4\}$ stage, the spatial size of the feature map is reduced by a factor of 2, meanwhile, the feature dimension increases by a factor of 2, which is different from the first stage mentioned above. The output of the last 3 stages forms the multi-scale feature maps, which are then fed to the neck. 

Figure \ref{coarse-to-fine}(b) shows the illustration of a single DCFT attention block, considering that the visual dependencies between the close regions are usually stronger than the far region, we maintain fine-grained attention to the surrounding pixels closest to the query tokens in the feature map and captures coarse-grained attention on distant regions after being pooling, so this attention mechanism has the far-to-close, coarse-to-fine character. With this character, we get a reasonable trade-off between computational overhead and long-range perception capabilities. 
  
As shown in Figure \ref{coarse-to-fine}(b), for a input feature map $z \in \mathbb{R}^{{H} \times {W} \times c} $, we first partition it into a grid of query windows with size $n_{wp} \times n_{wp}$, in which the query window shares the same surroundings. Each query window will pass through a linear projection layer to obtain the $Q_{i}=f_{q}\left(z^{1}\right)$ in the calculation of the attention mechanism, where $i$ indexes the $i$-th query window. For the $i$-th query window $Q_i \in \mathbb{R} ^{n_{wp} \times n_{wp} \times d}$, we split the input feature map $z$ into multiply sub-windows with the number of $N^{l} \times N^{l}$. Then we perform pooling for each feature map level $l \in \{1, ..., L\}$ to generate fine-grained and coarse-grained information. Since PMMW security images bear much more random noise than optical images, we adopt a denoising kernel in the pooling operation, with it has been shown in \cite{lopez2017deep} that denoising pre-processing is beneficial for object detection in PMMW security screening. A simple linear layer $f_{p}^{l}$ to pool the sub-windows spatially by:

\begin{equation} 
\label{eq1}
\begin{split}
& z^{l}=f_{p}^{l}(\hat{z}) \in \mathbb{R}^{\frac{H}{n^{l}} \times \frac{W}{n^{l}} \times c},\\
&\hat{z}=\operatorname{Reshape}(z) \in \mathbb{R}^{\left(\frac{H}{n^{l}} \times \frac{W}{n^{l}} \times c\right) \times\left(n^{l} \times n^{l}\right)},
\end{split}
\end{equation}
where $n^{l} =  \{n^{1} , \ldots, n^{L}\} $ is the size of sub-windows, and $n^{1}$ of the first feature map level is set to 1 to extract the finest granularity local feature. Once we obtain the pooled feature maps  $\{z^{1}, ..., z^{L}\}$ at all $L$ levels, we compute the key and value for all levels using two linear projection layers $f_k$ , $f_v$, i.e.  $K_i=f_{k}\left(\left\{z^{1},\ldots, z^{L}\right\}\right), V_i=f_{v}\left(\left\{z^{1}, \ldots, z^{L}\right\}\right)$. To perform DCFT attention, we need to first extract the surrounding tokens for each query token in the feature map. As we mentioned earlier, query tokens inside a query window $n_{wp} \times n_{wp}$ share the same surroundings. For the queries inside the $i$-th query window $Q_i \in \mathbb{R} ^{n_{wp} \times n_{wp} \times d}$, we extract the $N^{l} \times N^{l}$ keys and values from $K^{l}$ and $V^{l}$ around the window where the query lies in, and then gather the keys and values from all $L$ to obtain $ K_i = \{K_i^{1} , ...,K_i^{L}\} \in \mathbb{R}^{N \times c}$ and $ V_i = \{V_i^{1} , ..., V_i^{L}\} \in \mathbb{R}^{N \times c}$, where $N$ is the sum of DCFT region from all levels, i.e., $N=\sum_{l=1}^{L}\left(N^{l}\right)^{2}$. Finally, we follow \cite{liu2021swin} to introduce a relative position bias and compute the DCFT attention for $Q_i$ by :

\begin{equation} 
\label{eq2}
\begin{split}
\operatorname{Attention}\left(Q_{i}, K_{i}, V_{i}\right)=\operatorname{Softmax}\left(\frac{Q_{i} K_{i}^{T}}{\sqrt{c}}+B\right) V_{i},
\end{split}
\end{equation}
where $B =  \{B^{1} , \ldots, B^{L}\}$ is the learnable relative position bias. Similar to \cite{liu2021swin}, we parameterize the bias $B^{1} \in \mathbb{R}^{\left(2 n_{wp}-1\right) \times\left(2 n_{wp}-1\right)}$ in the first level, since the relative position along the horizontal and vertical axis are both lies in $[-n_{wp} + 1, n_{wp} - 1]$. After obtaining the attention scores of the feature map, we send the scores to the LayerNorm and Multi-Layer Perceptron (MLP) blocks.

The overall computational cost of our DCFT attention becomes $\mathcal{O} \left( \left(L + {\textstyle \sum_{l}} \left(n^{l}\right) ^{2}\right) (H W ) c \right)$, while the computational cost of original ViT and Swin Transformer are $\mathcal{O} \left( \left(4c + 2 H W\right) (H W ) c \right)$ and $\mathcal{O} \left( \left(4 c + 2 n_{wp}^{2}\right) (H W ) c \right)$, where $n_{wp}$ is the window partition size of Swin Transformer, which is usually set to 7. It can be seen that our proposed DCFT attention is more accurate with less computational complexity, and the accuracy comparison experiments are given in the ablation study in Section \ref{sec_ablation}.

\subsection{Multi-Scale Deformable Attention Neck}

Since PMMW security images contain high noise and sparse texture, the object and noise are relatively similar in the low-level feature dimension, which leads to network confusion and convergence difficulties. Therefore, a neck to further learn the high-dimensional semantic features and aggregate different classes of features is highly desired. DETR has been very successful in this direction. DETR proposes the idea of treating the object detection task as a set prediction to eliminate post-processing (e.g., NMS) while extracting the semantic features of individual instances well. The shortcomings of DETR are the slow convergence rate and the data-hungry characteristics\cite{wang2022towards}, which have been intensified in the low-resolution PMMW image situation. Therefore, the deformable attention module\cite{zhu2020deformable} is proposed, which has proven that its data-efficient properties\cite{wang2022towards} alleviate the inherent data-hungry problem of the DETR model. We follow the design of Deformabe DETR using the deformable attention module for further feature aggregation of the multi-scale feature maps output by the backbone.

Also, some recent work has demonstrated that the encoder of DETR is functionally identical to the region proposal network in traditional two-stage networks\cite{zhu2020deformable,meng2021conditional}, and the queries output by the encoder can be regarded as proposals in the region proposals network. Inspired by these works, we propose a method for introducing a spatial prior adapted to security screening, called query selection.

\begin{figure*}[htp]
\centering
\includegraphics[width=13cm]{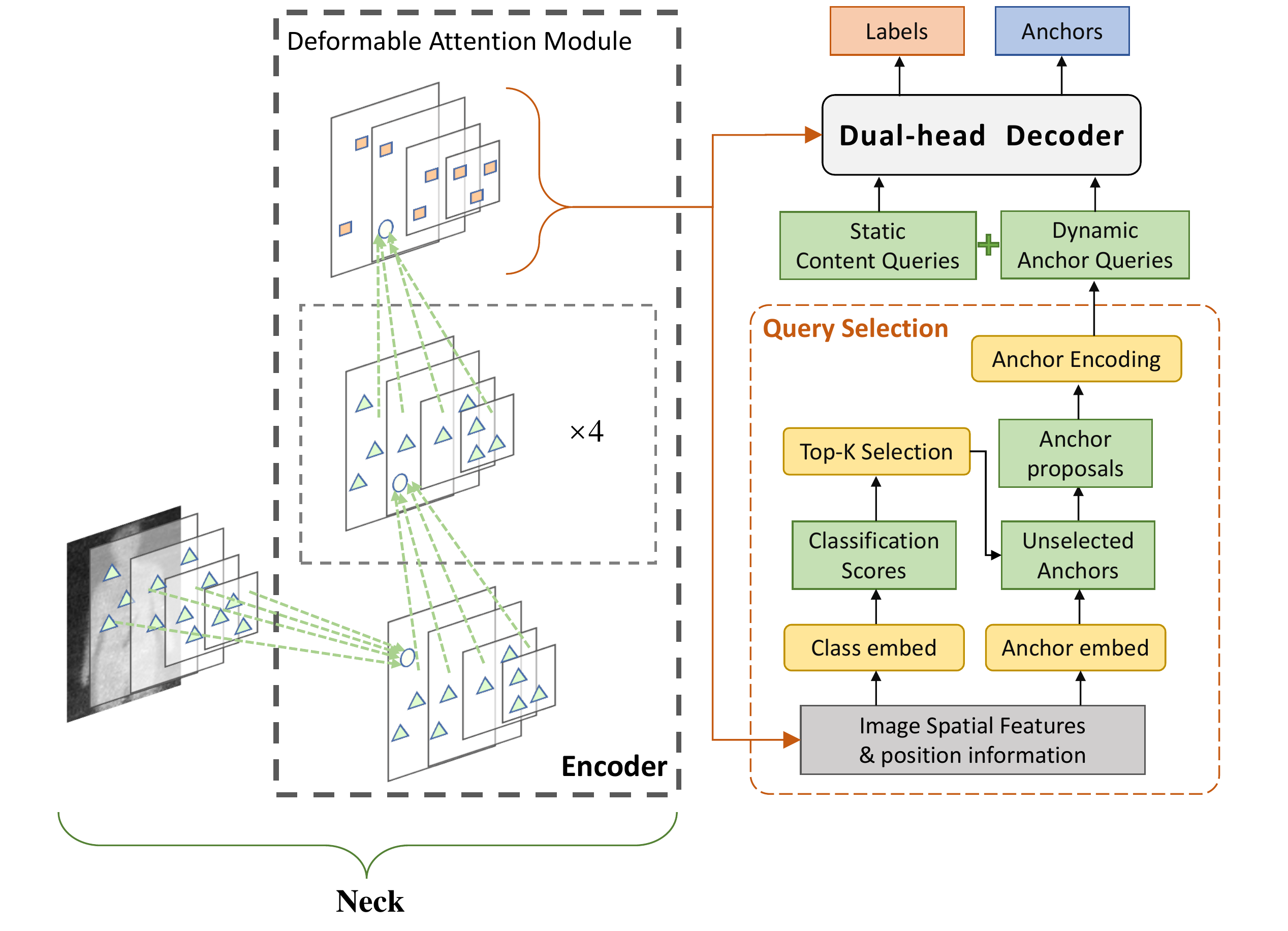}
\caption{Schematic of deformable attention neck. The Multi-Scale feature maps of the encoder output are used for the query selection and the cross-attention in the decoder.}
\label{neck}
\end{figure*}

 \subsubsection{Deformable Attention Module}
 
The deformable attention module only attends to a small set of key sampling points around a reference point, regardless of the spatial size of the feature maps, as shown in Fig \ref{neck}. We first take the output multi-scale feature map $\{{z^{s} \}}_{s=1}^{S} \in \mathbb{R}^{H \times W \times c}$  from the backbone, content feature  $z_{q} \in \mathbb{R}^{C}$, and its correspondent normalized coordinates of the 2-d reference point $\hat{\boldsymbol{p}}_{q} \in [0, 1]^{2}$ as input of deformable attention, where $q$ indexes a query element, $s$ indexes the input feature maps level. Then via two independent linear projection layers,  we use the query feature $z_q$ to obtain sampling offset $\Delta \boldsymbol{p} _{msqk} $ and attention weight $A_{msqk}$, where $m$ indexes the attention head, and $k$ indexes the sampling point and $\sum_{s=1}^{S} \sum_{k=1}^{K} A_{m s q k}=1$. The calculation of the multi-scale deformable attention can be expressed as: 

\begin{equation} 
\label{eq4}
\begin{aligned}
\operatorname{MS}&\operatorname{DeformAttn}\left(\boldsymbol{z}_{q}, \hat{\boldsymbol{p}}_{q},\left\{\boldsymbol{z}^{s}\right\}_{s=1}^{S}\right)=\sum_{m=1}^{M} \boldsymbol{W}_{m}\cdot \\
&\left[\sum_{s=1}^{S} \sum_{k=1}^{K} A_{m s q k} \cdot \boldsymbol{W}_{m}^{\prime} \boldsymbol{z}^{s}\left(\phi_{s}\left(\hat{\boldsymbol{p}}_{q}\right)+\Delta \boldsymbol{p}_{m s q k}\right)\right],
\end{aligned}
\end{equation}
where $\phi_{s} \left(\hat{\boldsymbol{p}}_{q} \right)$ re-scales the normalized coordinates $\hat{\boldsymbol{p}}_{q}$ to the input feature map of the $s$-th level, $\boldsymbol{W}_{m}^{\prime} \in \mathbb{R}^{C_{v} \times C} $and $\boldsymbol{W}_{m} \in \mathbb{R}^{C \times C_{v}}$ are of learnable weights ($C_{v} = C/M$).

\subsubsection{Query Selection Module} 

As shown on the right side of Figure \ref{neck}, the object queries used as input to the decoder contain spatial queries and content queries\cite{meng2021conditional, liu2022dab}. In DETR \cite{carion2020end}, both queries are randomly initialized as static embeddings without using any encoder features as spatial prior. For the spatial queries, DAB-DETR\cite{liu2022dab} has proved that anchor boxes are better queries for DETR, which learns anchor boxes directly instead of learning the spatial queries. For the content queries, DETR set them as all zero vectors, while Deformable DETR\cite{zhu2020deformable} learns content queries together with the spatial queries, which is an alternative approach to implementing query initialization. Furthermore, Deformable DETR also has a similar query selection module that is called "two-stage", that is, both the spatial and content queries are generated by the top-K selected features. Similarly, Efficient DETR \cite{yao2021efficient} also selects top-K features based on the classification score of each image spatial feature. Furthermore, rather than the static 2d queries in DETR, the initialization of dynamic 4d anchor boxes in PMMW-DETR makes it more closely related to spatial queries so that it can be spatial prior to being introduced by query selection.

Inspired by the above practice\cite{zhu2020deformable,carion2020end,yao2021efficient}, we propose a query selection module that combines the static content queries initialization and the dynamic anchor queries selection. Dynamic anchor queries are helpful to enhance spatial queries. However, since the multi-scale features contain rich content information, we make content queries static and learnable to avoid introducing incomplete objects that may lose information and mislead the decoder. To give more detail, as shown in Figure \ref{neck}, the output multi-scale features are first embedded into classification scores and anchors by the class embedding layer and anchor embedding layer, respectively. Then we select the top-K anchors as proposals based on the classification score. Finally, we obtain dynamic anchor queries by encoding the selected anchor proposals while initializing the static content queries as original DETR. 
 
By the query selection module, the decoder of PMMW-DETR learns the spatial distribution information directly from the feature maps output by the encoder, instead of starting from zero with blind queries. \cite{zhu2020deformable} has shown that object queries will query the object in its spatial distribution range, and the human security images happen to all have a human body as the subject with high similarity. Besides, as mentioned above, in order to avoid misleading the network, we refine the content queries layer by layer. The detailed comparative ablation experiments are given in Section \ref{sec_ablation}.
 
\subsection{Task-Aligned Dual-Head Block}

\begin{figure*}[htp]
\centering
\includegraphics[width=13cm]{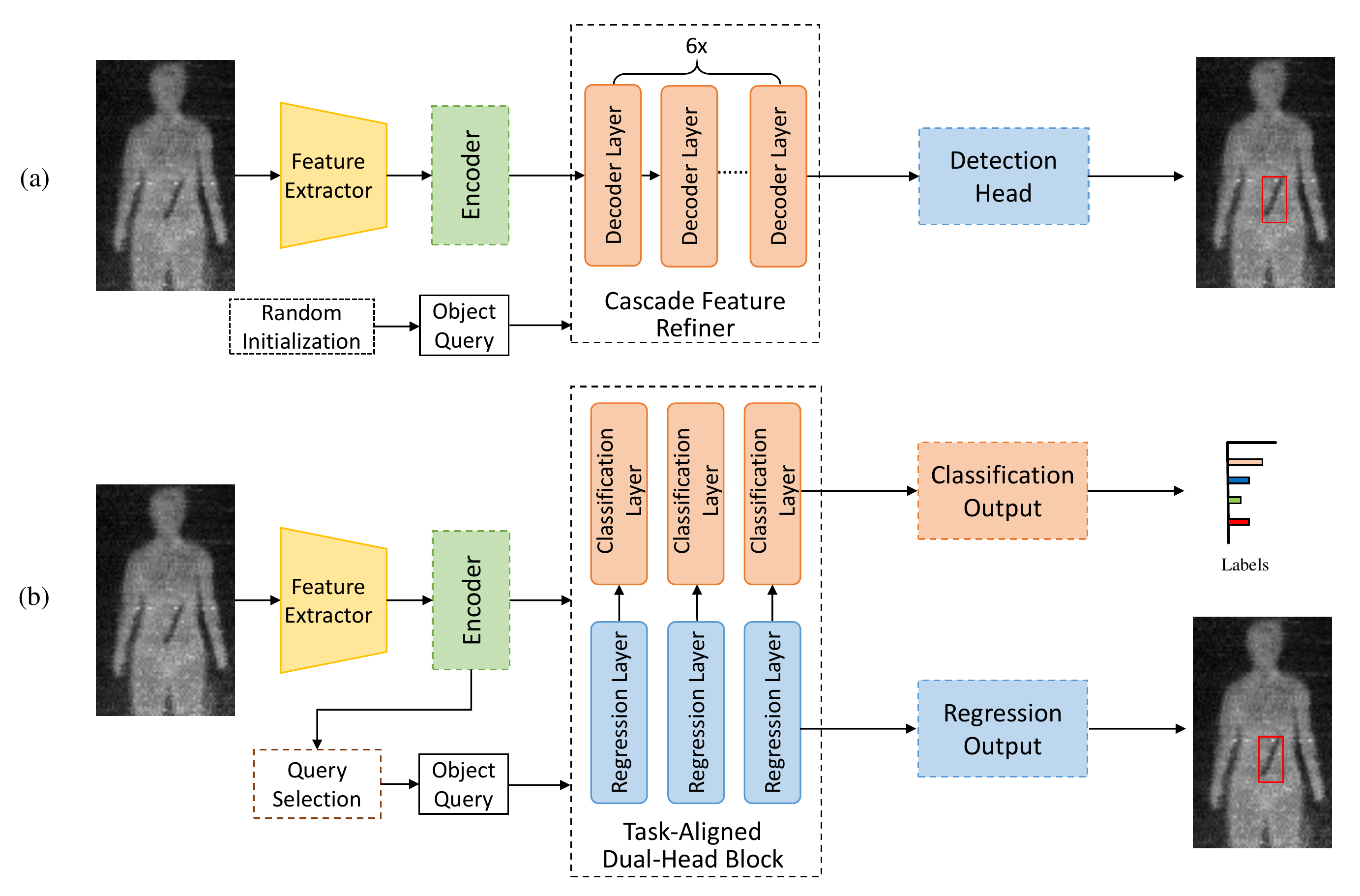}
\caption{Previous end-to-end detectors like DETR and Deformable DETR (a) and ours (b).}
\label{dual_head}
\end{figure*}

Most current concealed object detection networks share a head for both classification and bounding box regression. However, \cite{wu2020rethinking, song2020revisiting, feng2021tood} have proved that there is a spatial misalignment between classification and regression tasks. This misalignment can be explained from two aspect. In the one hand, the two tasks have different sensitive locations for the same object. For example, some salient areas may be beneficial for classification while the boundary might have rich information for regression. The spatial task misalignment greatly limits the performance of concealed object detection. Therefore, we propose a task-aligned dual-head block. We perform concealed object classification and regression independently by using two independent branches in parallel heads. But such a two-branch design might lead to a lack of interaction between the two tasks, resulting in inconsistent predictions when performing them, \textit{i.e. } two different decoders focus on regions that are too far apart from each other, which creates a spatial misalignment on the other hand. So these two problems can be jointly modeled as a trade-off between task decoupling and spatial consistency of two parallel branches. To avoid this problem, we propose query sharing to explicitly align the two tasks with the sharing query cross attention.

As shown in Figure \ref{dual_head}, the proposed task-aligned dual-head block has two different types of decoder layers with query sharing modules. Different from the six-layer structure of the original DETR decoder, we design a decoder layer with two branches and add the decoder to the head to refine the features. The dual-head design achieves decoupling the classification and regression tasks, avoiding the inherent conflict between these two tasks. The regression head consists of regression layers on the left, and on the right side is the classification head consisting of classification layers. The regression layers and classification layers are three layers each, and the total number of layers is 6, which is consistent with the original DETR. In the PMMW-DETR network, the regression layer is responsible for refining the anchor boxes, while the classification layer refines the class labels. Therefore, in the following, we call the regression layer the anchor refinement layer and the classification layer the class refinement layer. 

\begin{figure*}[htp]
\centering
\includegraphics[width=15cm]{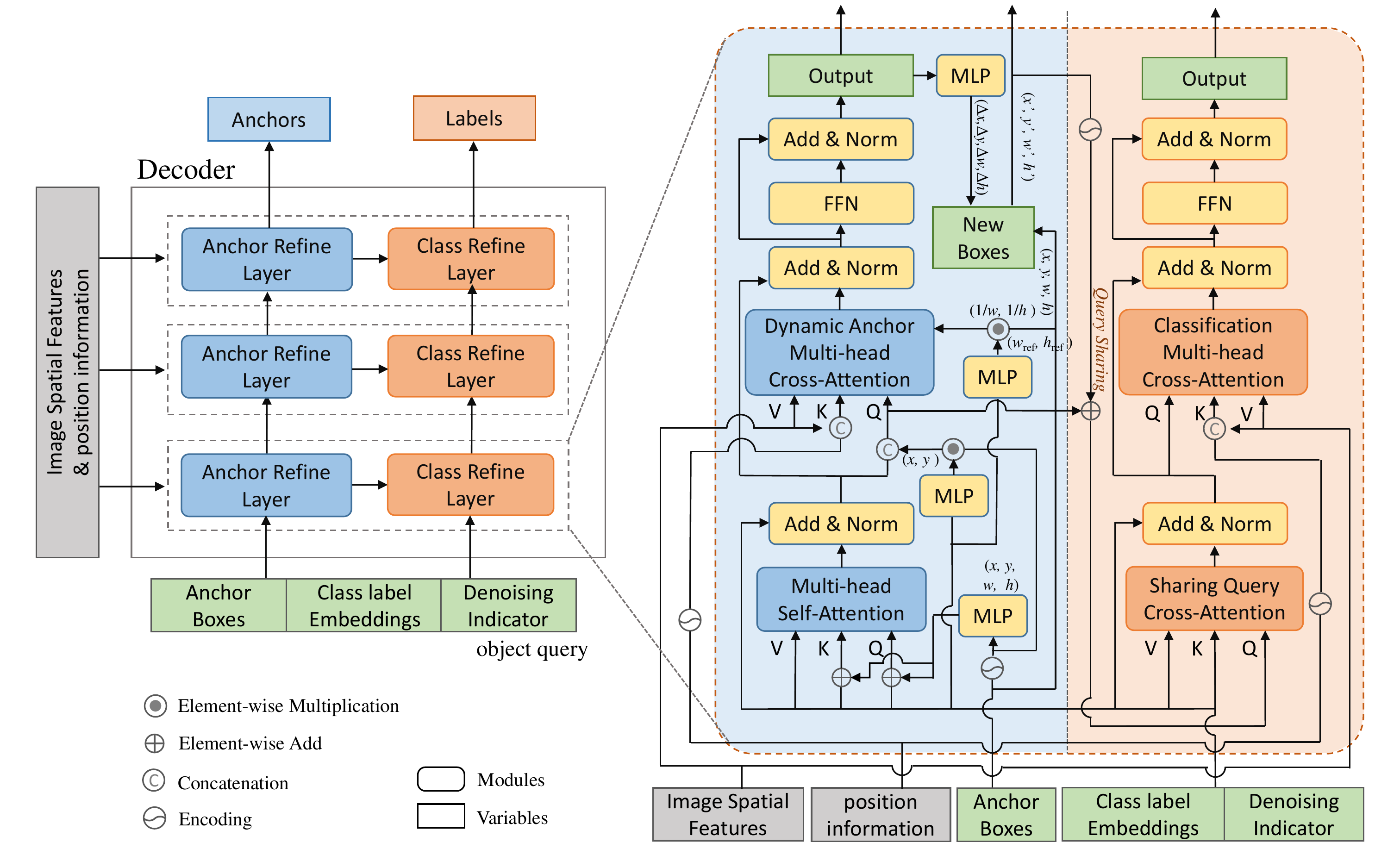}
\caption{The illustration of the task-aligned dual-head block. Three task-aligned dual-headed blocks consist of the decoder.}
\label{head_illustration}
\end{figure*}

Figure \ref{head_illustration} shows the specific data flow of the task-aligned dual-head block. Each block layer includes a self-attention module and a cross-attention module. The self-attention module is used for query updating, and can be expressed as:
\begin{equation}
\label{eq-self-att-eq}
\begin{split}
\operatorname{SelfAttn}(Q, K, V)=\operatorname{softmax}\left(\frac{Q K^{T}}{\sqrt{d}}\right) V,
\end{split}
\end{equation}
where $d$ is the number of queries, and queries $Q_{q}$, keys $K_{q}$, and values $V_{q}$ can be expressed as
\begin{equation}
\label{eq-self-att}
\begin{split}
Q_{q}=C_{q}+P_{q}, \quad K_{q}=C_{q}+P_{q}, \quad V_{q}=C_{q},
\end{split}
\end{equation}
where $C_{q} \in \mathbb{R}^{D}$ indicates the learnable content query, and $P_{q} \in \mathbb{R}^{D}$ indicates the spatial query generated by
\begin{equation}
\label{eq-embed-P}
\begin{split}
P_{q}=\operatorname{MLP}\left\{\operatorname{Cat}\left[\operatorname{PE}\left(x_{q}\right), \operatorname{PE}\left(y_{q}\right), \operatorname{PE}\left(w_{q}\right), \operatorname{PE}\left(h_{q}\right)\right]\right\},
\end{split}
\end{equation}
where $(x_q,y_q,w_q,h_q)$ is the $q$-th anchor generated from the query selection module; $\operatorname{PE}: \mathbb{R} \rightarrow \mathbb{R}^{D / 2}$ is the positional encoding function to generate sinusoidal embeddings; the notion Cat means concatenation function; $\operatorname{MLP}: \mathbb{R}^{2D} \rightarrow \mathbb{R}^{D}$ is composed of two linear layers.

The cross-attention module is used for feature probing, where queries, keys, and values can be expressed as
\begin{equation}
\label{eq-cross-att}
\begin{split}
&Q_{q}=\operatorname{Cat}\left(C_{q}, \operatorname{PE}\left(x_{q}, y_{q}\right) \cdot \operatorname{MLP}\left(C_{q}\right)\right), \\ & K_{x, y}=\operatorname{Cat}\left(F_{x, y}, \operatorname{PE}(x, y)\right),\\ & V_{x, y}=F_{x, y},
\end{split}
\end{equation}
where $\operatorname{MLP}: \mathbb{R}^{D} \rightarrow \mathbb{R}^{D}$ is composed of two linear layers to learn a scale vector of the content information. We concatenate the position and content information together as queries and keys to consider both the content and position contributions so that the query matrix and key matrix can be expressed as $Q=\operatorname{Cat}(Q_C, Q_P)$ and $K=\operatorname{Cat}(K_C, K_P)$. Based on the above queries, keys, and values, the cross-attention module is formulated as follows:
\begin{equation}
\label{eq-cross-att-eq}
\begin{aligned}
\operatorname{Cross}&\operatorname{Attn}(Q, K, V)=\\ &\operatorname{softmax}\left(\frac{Q_CK_C^T+\operatorname{ModAttn}(Q_P,K_P)}{\sqrt{d}}\right) V,
\end{aligned}
\end{equation}
where the modulated positional attention helps us extract features of objects with different widths and heights, which can be expressed as
\begin{equation}
\label{eq-modulateAttn}
\begin{aligned}
&\operatorname{ModAttn}(\operatorname{PE}(x_q,y_q),\operatorname{PE}(x,y))\\
&=\left(\operatorname{PE}\left(x_{q}\right) \operatorname{PE}\left(x\right)^T\frac{w_{\mathrm{ref}}}{w}+\operatorname{PE}\left(y_{q}\right) \operatorname{PE}\left(y\right)^T\frac{h_{\mathrm{ref}}}{h} \right)/\sqrt{D},
\end{aligned}
\end{equation}
where $1/\sqrt{D}$ is used for value rescaling \cite{vaswani2017attention}, and the reference width and height that are calculated by $w_{\mathrm{ref}}, h_{ \mathrm{ref}}=\sigma\left(\operatorname{MLP}\left(C_{q}\right)\right)$ as shown in Figure \ref{head_illustration}.

The two content queries and spatial queries are updated layer by layer. Using coordinates as spatial queries for learning makes it have clear spatial meaning. As shown in Figure \ref{head_illustration}, each anchor refines layer outputs an updated object anchor by predicting the relative positions $(\Delta x,\Delta y, \Delta w, \Delta h)$. For the sake of decoupling the classification and regression tasks, we only consider the output of the class refine layer as the classification result. Further, we design a query sharing mechanism to enhance the collaboration between these two tasks, that is, the updated anchor of the anchor refine layer is used as the input spatial query of the sharing query cross attention in the class refine layer after being embedded by Eq.\ref{eq-embed-P}. The calculation method of the sharing query cross attention is simple, which is the same as Eq.\ref{eq-self-att-eq}. However, the sharing query cross attention effectively mitigates the problem of inter-spatial variation by exploiting the property of attention aggregation target correlation, and further enhances the spatial consistency of the two branches, completing the task-aligned dual-head block. The validity of the query sharing is verified in Section \ref{sec_ablation}. In addition, we introduce the query denoising learning\cite{li2022dn}, which adds a denoising part containing a denoising indicator and denoising loss as a training shortcut to accelerate the training of PMMW-DETR.


\section{Experiments and Results}
\subsection{Experimental Environments}

\begin{figure}[htp]
\centering
\includegraphics[width=8.5cm]{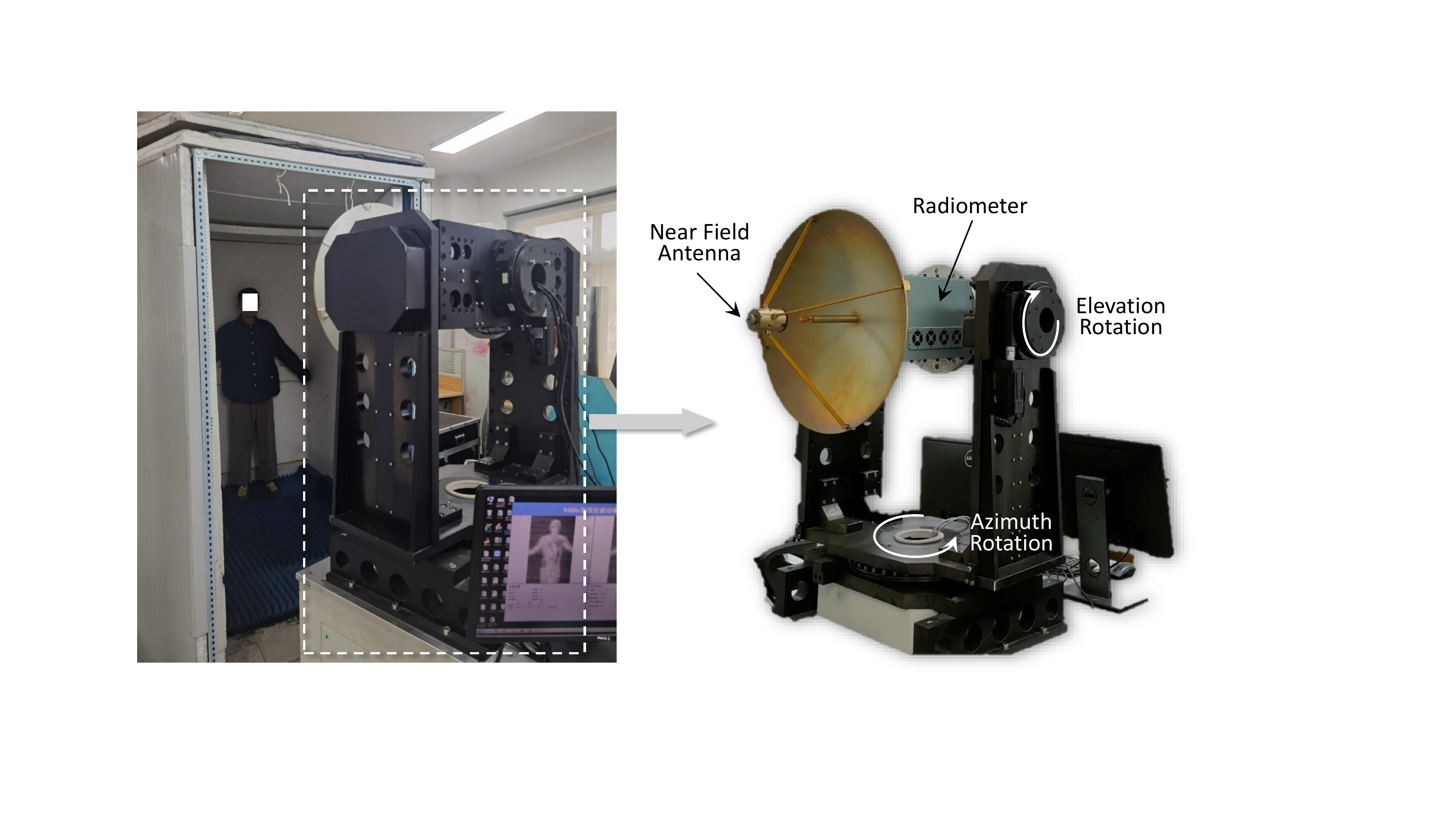}
\caption{The PMMW security screening system and Experimental Environments.}
\label{Sec_Experimental_equipment}
\end{figure}

The dataset was acquired by our self-developed PMMW security system, as shown in Figure \ref{Sec_Experimental_equipment}. The system consists of a Cassegrain antenna, a radiometer consisting of an ortho-mode transducer, two direct detection modules, a data acquisition module, and a three-axis scanning turntable. The system operates in the 94±2 GHz band with a sensitivity of about 0.4 K. Using an antenna with a diameter of 0.6096 m, it can achieve a resolution of 0.36°. As a research-oriented rather than a commercial imaging system, the system uses a scanning imaging regime that trades a longer imaging time (approximately 5 min per image) for better imaging quality. Compared with existing publicly available datasets \cite{lopez2017deep,lopez2018using,pang2020real,yang2022transformer}, the image quality of our dataset is closer to or even surpasses that of reported commercial security systems, such as X250 and S350 of Millivision Inc., iPat imager of Trex Enterprises Inc., and SP0-NX of Qinetiq Inc. So our dataset is more representative and forward-looking than the existing datasets. 

\subsection{Security Dataset and Implementation Details}
\label{Sec_dataset_and_detail}

We collected a total of 247 security images, in which the types of concealed objects included the metal wrench, alcohol bottle, metal knife, and metal pistol. 186 PMMW security images formed the training set and the remaining 77 formed the test set. The training of the Transformer network requires a large amount of data, but the scanning imaging regime is too time-consuming to form a large-scale dataset. To compensate for the lack of data set, we added 379 simulated images to the original training set. The simulation image is implemented based on the ray tracing method \cite{salmon2020indoor,qi2016passive}. To sum, the above original dataset has a total of 715 images, of which 638 images form the training set and 77 images form the test set. After a series of data augmentation operations such as flipping, rotating, Gaussian degradation filtering, and changing the brightness, the training set can be expanded by a factor of seven to 4466 images. All our datasets will be available at https://github.com/Ch3ngguo/opening-source-PMMW-dataset.

\begin{figure}[htp]
\centering
\includegraphics[width=8.8cm]{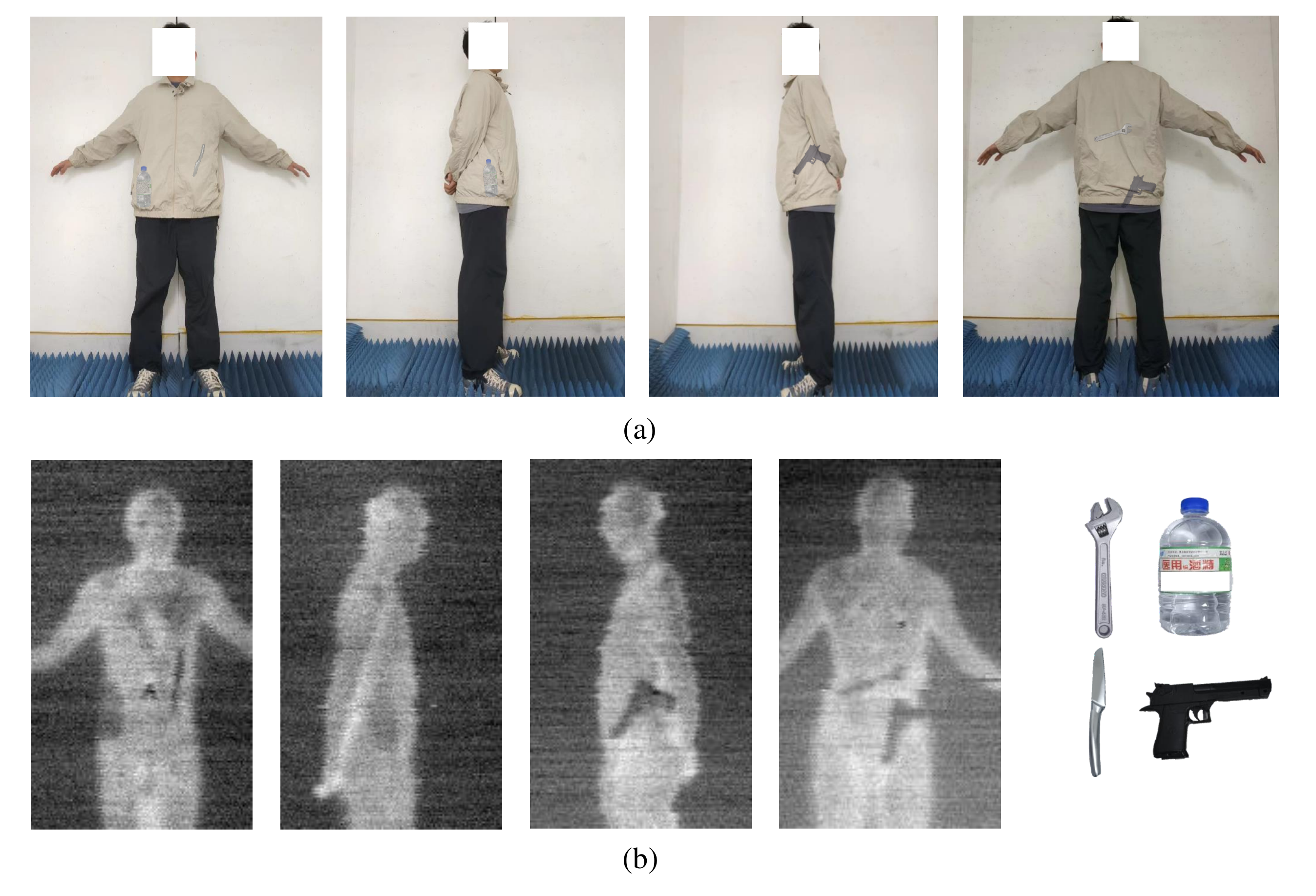}
\caption{(a) Optical schematic photos corresponding to human security screening. (b) Representative PMMW security images and the hidden objects we used, which include a metal wrench, alcohol bottle, metal knife, and metal pistol.}
\label{Experi_human}
\end{figure}

Several typical PMMW security images are given in Figure \ref{Experi_human}, from which it can be seen as follows:

1) Due to the uneven ambient brightness, the human chest resembles the texture features of the alcohol bottle, which may lead to false alarms.

2) The metal wrench in the fourth image is thin and shares similar contour features with the scanning noise, and the brightness of the alcohol bottle in the second image is close to the human body, both of which may lead to missed detection.

3) Since the flat place of the human body will reflect the strong brightness, such as the central part in the first image, which may affect the detection of the central hidden objects.

Therefore, it is a difficult task to detect objects on such a low resolution, high noise, and low little texture information dataset.

For the implementation details, we use {1, 3, 5, 7} as the size of $n^{l}$ in the DCFT module. We add uniform noise on boxes and set the hyperparameters with respect to noise as $\lambda_{1} = 0.4$, $\lambda_{2} = 0.4$, and $\gamma = 0.4$ in the query denoising learning. We train the network for 50 epochs and initialized parameters by Xavier. We adopt AdamW \cite{loshchilov2017decoupled} as the optimizer with a weight decay of $1 \times 10^{-4}$. The initial learning rate is $1 \times 10^{-4}$ and it drops by multiplying 0.1 at the 40-th epoch. The batch size is 4, and all the models are trained on a single NVIDIA TESLA A30 GPU.

\subsection{Comparison with the State-of-the-Art}

Following the mainstream approach of object detection evaluation, we use average precision (AP) and average recall (AR) to evaluate the performance of the proposed model. In order to show the detection results for all classes, the mAP/mAR is obtained by summing the AP/AR of each class and taking the average. Essentially, AR is twice the area under the recall-IOU curve, AP is the area under the precision-recall curve (PR curve), where recall is the ratio of detected samples to the actual total samples, and precision reflects the detection rate. The calculation of mAP and mAR is given as follows:

\begin{equation}
\label{mAP}
\begin{aligned}
\text { mAP } =\frac{\sum_{k=1}^{K} \mathrm{AP}_{k}}{K}, \quad\text { mAR } =\frac{\sum_{k=1}^{K} \mathrm{AR}_{k}}{K},
\end{aligned}
\end{equation}
where $K$ is the number of classes, the AP$_{k}$ and AR$_{k}$ are given as follows:

\begin{equation}
\label{AP}
\begin{aligned}
&\text { AP$_{k}$ } =\sum_{i=1}^{n-1}\left(r_{i+1}-r_{i}\right) p_{\text{interp}}\left(r_{i+1}\right),\\
&{\text { AR$_{k}$ }=\frac{1}{O} \sum_{o=1}^{O} \max \left\{\text{Recall}(\mathrm{c} ,t(o))\right\},}
\end{aligned}
\end{equation}
where $t(\textit{o})$ is a set of IoU thresholds and $\mathrm{c}$ is the confidence values of all the detection results, $r_{1}, r_{2}, \ldots, r_{n}$ are the recall levels (in ascending order) at which the precision is first interpolated. The interpolated precision $p_{\text {interp }}(r)=\max _{r^{\prime} \geq r} \text {Precision} \left(r^{\prime}\right)$, defined as the highest precision found at a certain recall level $r$, for any recall level $r^{\prime} \geq r$. The Precision and Recall are defined as:

\begin{equation}
\label{Precision}
\begin{aligned}
\text {Precision} =\frac{\mathrm{TP}}{\mathrm{TP}+\mathrm{FP}},\quad \text {Recall} =\frac{\mathrm{TP}}{\mathrm{TP}+\mathrm{FN}},
\end{aligned}
\end{equation}
where the TP is true positives, FN is false negatives, and FP is false positives.
\begingroup
\setlength{\tabcolsep}{6pt} 
\renewcommand{\arraystretch}{1} 
\begin{table*}[]
\renewcommand{\arraystretch}{1}
\centering
\caption{\bf Comparison with the SOTA object detection models.}
\label{SOTA}
\renewcommand{\arraystretch}{1.2}
\setlength\tabcolsep{2mm}{
\begin{tabular}{lcccccccc}
\hline
\hline
Method           & Epochs & mAP$_{50}$ & mAP$_{75}$ & mAP  & mAP$_{S}$  & mAP$_{M}$  & AR$_{100}$ & End-to-End\\
\hline
\textit{One-stage model} \\

YOLOv3           &273     & 95.0     & 76.4     & 61.0      & 39.3      & 62.8      & 65.2      &       \\
SSD              &24      & 95.5     & 65.3     & 59.5      & 31.4      & 62.7      & 64.3      &       \\
\hline
\textit{Two-stage model} \\

Faster-RCNN      &24      & 95.7     & 55.7     & 56.1      & 31.9      & 58.4      & 61.0      &      \\
\hline
\textit{Anchor-free model} \\

ATSS             &24      & 96.2     & 74.1     & 59.8      & 32.2      & 61.5      & 64.0      &      \\
FCOS             &24      & 95.7     & 60.0     & 58.8      & 34.4      & 60.1      & 62.1      &       \\
\hline
\textit{Transformer-based model} \\

PVT              &24      & 96.1     & 80.4     & 64.0      & 39.8      & 65.3      & 69.1      &      \\
Dyhead           &24      & 96.3     & 74.0     & 64.2      & 38.4      & 66.9      & 68.5      &  \\
\hline
\textit{Strong SOTA model} \\

YOLOv8  &500      & 95.6     & 83.4     & 65.6      & 45.5      & 67.6     & 69.8      &    \\
\hline
PMMW-DETR  &24      & 97.0     & 87.1     & 67.8      & 45.7      & 68.4      & 75.5      & \checkmark   \\
PMMW-DETR(Ours)        &50      & \textbf{97.5}     & \textbf{91.8}     & \textbf{71.3}      & \textbf{49.1}      & \textbf{74.1}      & \textbf{80.4}      & \checkmark       \\
 \hline
 \hline
\end{tabular}
}
\end{table*}
\endgroup
To show the superior performance of our proposed model, we present experimental results comparison with six representative state-of-the-art (SOTA) detectors, including YOLOv3 \cite{redmon2018yolov3} (one-stage detector), SSD \cite{liu2016ssd} (one-stage detector), Faster-RCNN \cite{ren2015faster} (two-stage detector), ATSS \cite{zhang2020bridging} (anchor-free detector), FCOS \cite{tian2019fcos} (anchor-free detector), PVT \cite{wang2021crossformer} (Transformer-based detector), and Dynamic Head (Transformer-based detector) \cite{dai2021dynamic}. In order to ensure the comprehensiveness of the comparison test, each of the six object detectors represents the different characteristics of object detectors in different periods. YOLOv3 and SSD are the most reliable one-stage detectors with the best small object detection ability in the industry and have recently been used to detect concealed objects \cite{pang2020real,kowalski2019real}, which we introduce to compare with the work of predecessors. Faster-RCNN is one of the most classical two-stage detectors. Faster-RCNN proposed to use Region Proposal Net (RPN) to obtain accurate anchors and then match objects, which has certain similarities with the decoder structure of DETR. ATSS and FCOS are the two most pioneering anchor-free algorithms. FCOS first proposed the concept of the center-ness and achieved adaptive detection by regressing the distance from the center point to the boxes. Since no prior anchor boxes are required, FCOS has strong robustness to the complex concealed object detection working environment. ATSS further proposed adaptive positive and negative sample selection on the basis of FCOS, which also inspired Yang \textit{et al.} \cite{yang2022transformer}. To demonstrate the superiority of our proposed attention computation method on the backbone and head, we compare our PMMW-DETR with PVT and Dyhead, which introduce the attention mechanism into the backbone and head networks, respectively. In addition, we added the latest YOLOv8 detector from the YOLO series as a strong SOTA model to make our SOTA comparison experiments more convincing. To ensure the fair validity of the experiment, we adopt grid search \cite{lavalle2004relationship} to find the relatively  optimal hyperparameters in the above networks, and the accuracy comparison experiments are given in Table \ref{hyperparameters_settings}. All the experiments were run on the same dataset, and the above models are implemented based on the mmyolo and mmdetection \cite{chen2019mmdetection}.

\begin{figure}[htp]
\centering
\includegraphics[width=8.8cm]{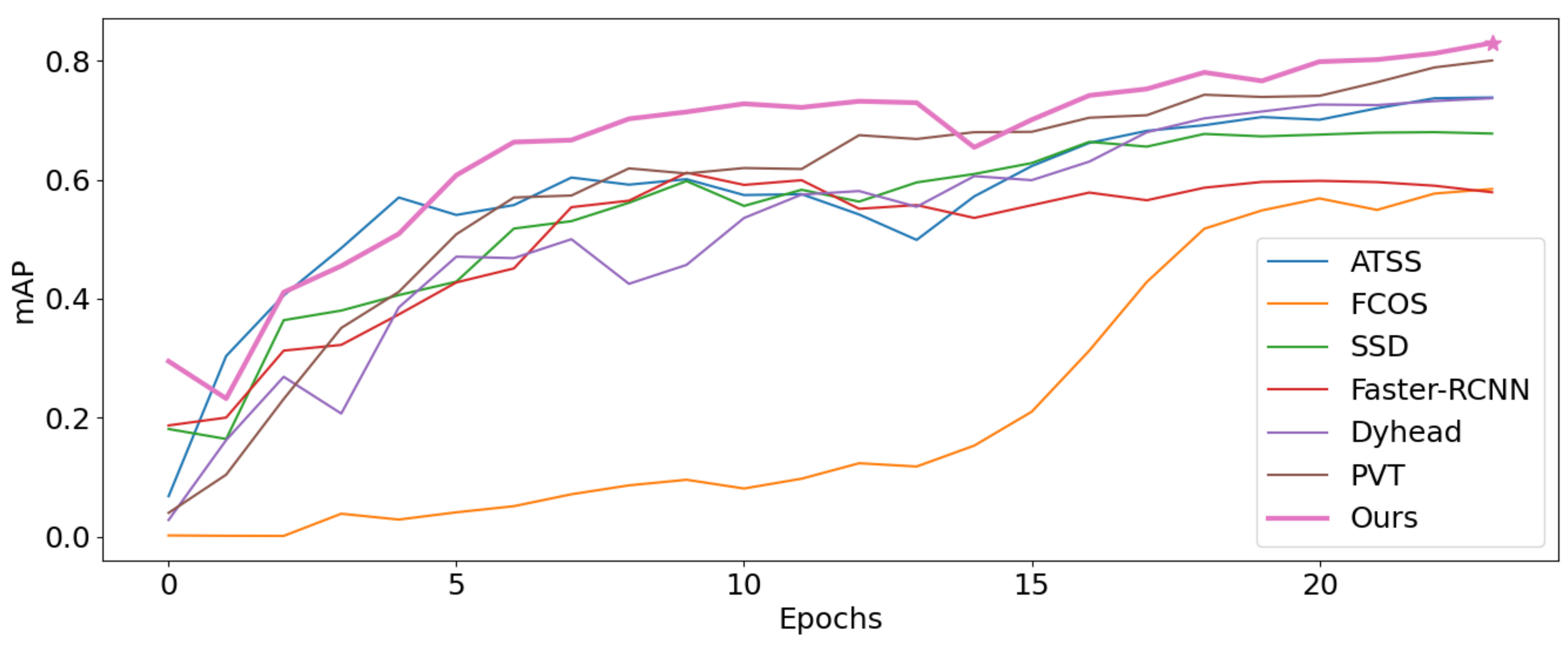}
\caption{The 24 epochs mAP$_{75}$ curves of PMMW-DETR and other models.}
\label{curve_SOTA}
\end{figure}
In the Table \ref{SOTA}, the epoch is the metric used to measure the convergence speed, where an epoch is a process of sending all data into the network and completing a forward calculation and back propagation. To fairly compare the performance, we set up an experiment with 24 (2x schedule) epochs setup and drop the learning rate of our model by multiplying by 0.1 in the 20-th epoch. The mAP$_{50}$ is the standard metric for Pascal VOC, which defines the mapped metric using a single IoU threshold of 0.5. Similarly, the mAP$_{75}$ uses a single IoU threshold of 0.75. The mAP is the average of mAP under 10 IoU thresholds (i.e.,$ 0.50, 0.55, 0.60, \ldots , 0.95$), which is the most strict metric in the COCO challenge (The mAP of the SOTA model in COCO dataset is 63.3). The mAR$_{100}$ means the average recall calculated when one hundred anchor boxes are given to the images. The End-to-End indicates whether the model does not require NMS and manual design anchors. 

Besides, Figure \ref{curve_SOTA} shows the performance curves of ours and other models. It can be seen that we are superior to other methods after the fifth epoch.

\begin{figure*}[htp]
\centering
\includegraphics[width=16cm]{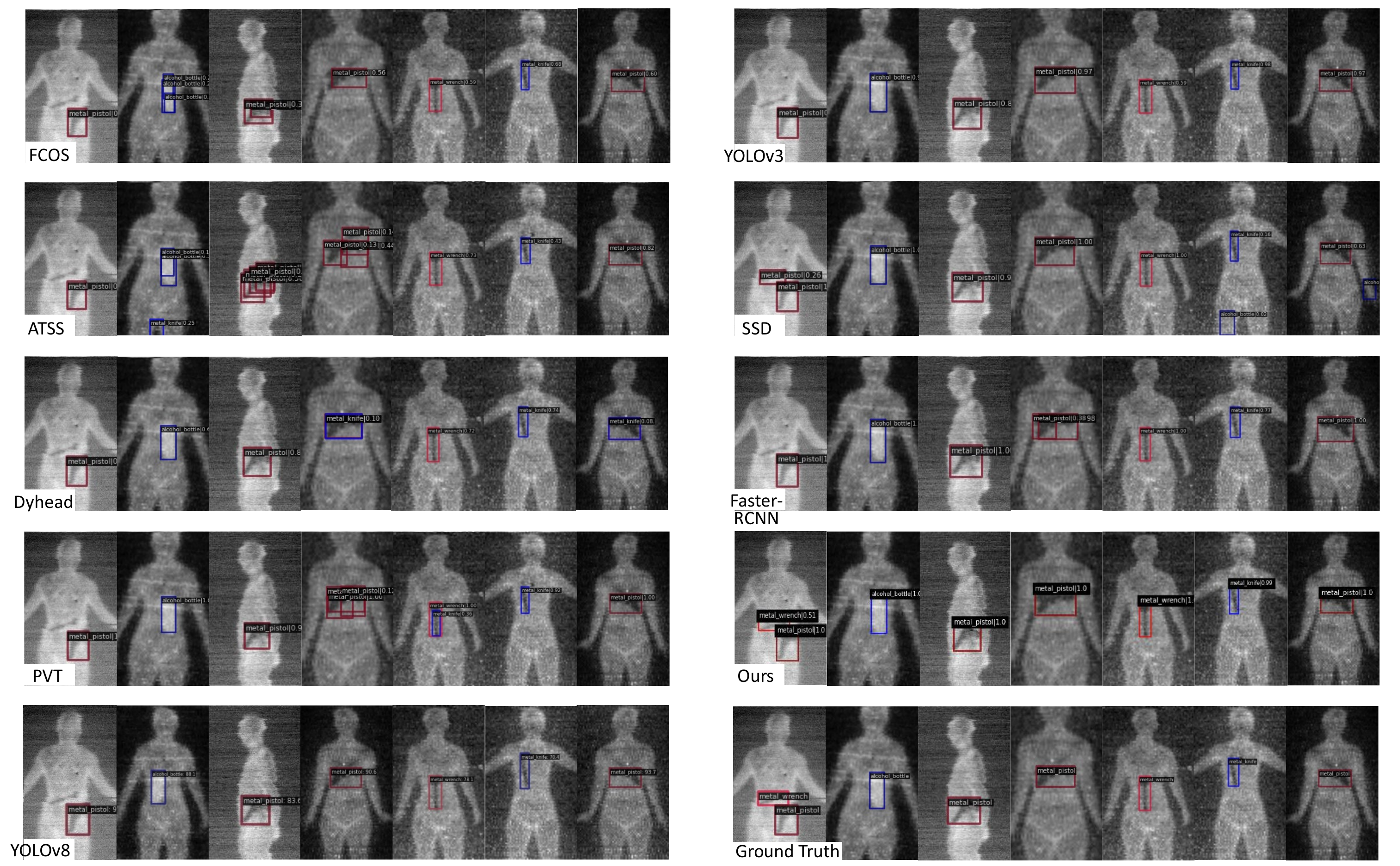}
\caption{Visualization detection results of SOTA models and our proposed model.}
\label{Detect_result}
\end{figure*}

\begin{figure}[htp]
\centering
\includegraphics[width=8cm]{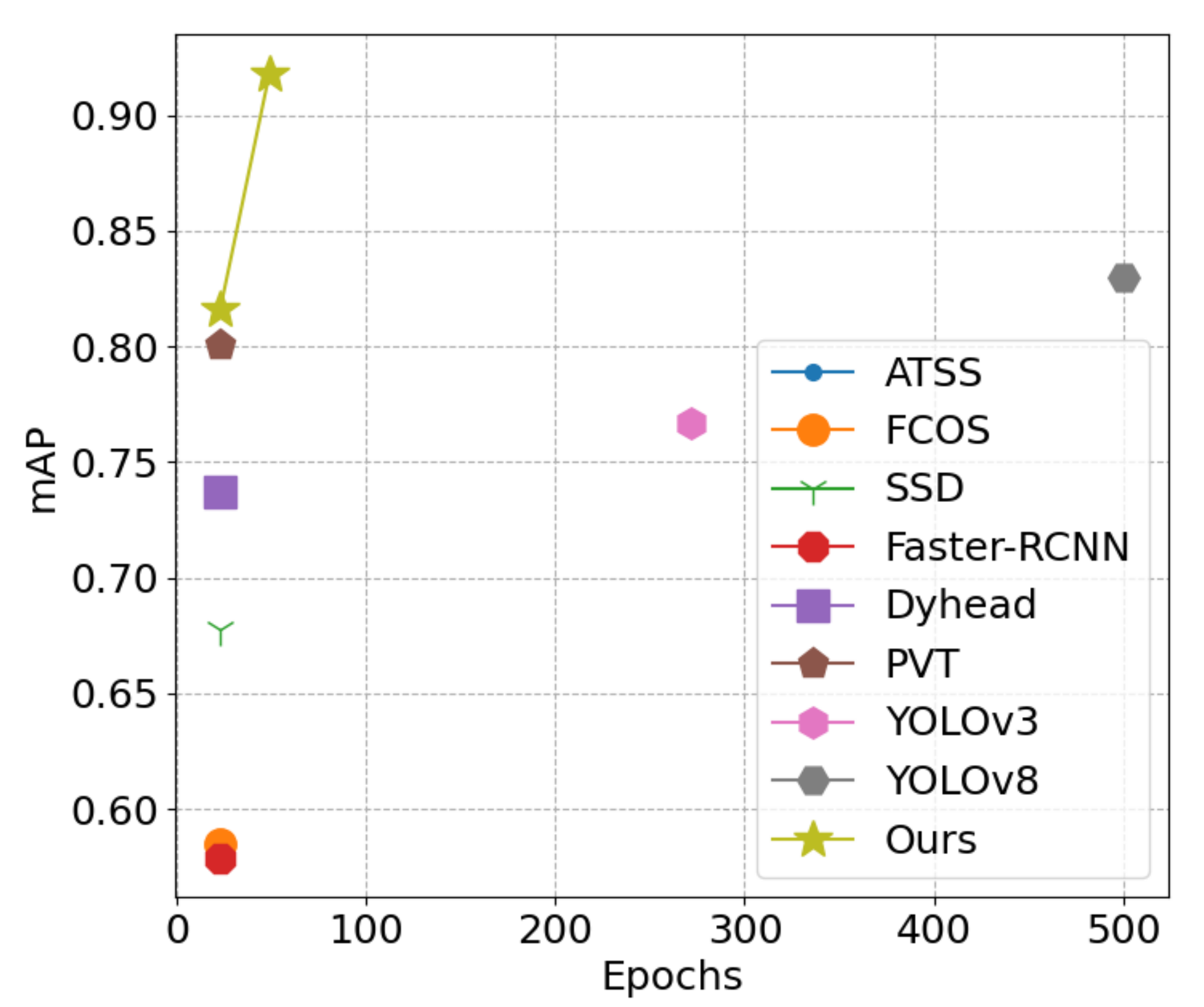}
\caption{The mAP$_{75}$ comparison of PMMW-DETR and other models.}
\label{points_SOTA}
\end{figure}

\textbf{YOLOv3} is a widely used one-stage detector, which was already adopted for PMMW security screening \cite{pang2020real}. It directly learns the bounding box, confidence, and category probability by regression. YOLOv3 has a fast detection speed and a low false detection rate because it can see global information. However, the detection accuracy of YOLOv3 is lower than other SOTA detectors, especially for small objects.
It can be seen that our method is +10.4\% mAP higher than YOLOv3. We believe that the low detection rate is closely related to the fact that the original YOLOv3 model does not include multi-scale detection. Besides, YOLOv3 often requires more than two hundred epochs to converge, which is unacceptable on PMMW datasets. \textbf{YOLOv8} is the latest SOTA model of the YOLO series. With the introduction of new technologies such as decoupling head and distribution focal loss, YOLOv8 achieves fair AP performance. Nevertheless, YOLOv8 have to trained with 500 epochs, even longer than YOLOv3.

\textbf{SSD} is another widely used one-stage detector It introduces the pyramid feature hierarchy, that is, objects are predicted on the feature maps of different receptive fields, thus improving the detection accuracy of the network. However, as shown in Table \ref{SOTA}, the $mAP_{S}$ of SSD for small objects is very low, only 31.4\% (18.9\% lower than our method). At the same time, there are missed detection and wrong classification as shown in Figure \ref{Detect_result}. As in the fifth figure, the confidence level of the metal knife is only 0.16, which will be a false negative object in practical applications.

\textbf{Faster-RCNN} is a classic two-stage detection network. It introduces RPN to generate anchor points and nine proposal anchors around each point. After screening, the effective proposal anchor is sent to the classification and regression network. As we can see from the Table \ref{SOTA}, our model is 36.1\% higher in $mAP_{75}$ and 19.4\% higher in AR than Faster-RCNN. It can be seen from Figure \ref{Detect_result} that Faster-RCNN has fewer boxes with higher confidence and thus is easier to miss objects.

The above methods need manually designed anchors as prior information for the network. On the contrary, \textbf{FCOS} is a famous anchor-free detector. It treats object detection as a regression of the distance between each position on the feature map and the bounding box. \textbf{FCOS} avoids all the super parameters related to the proposal anchor, so saves the memory occupation and improves the calculation efficiency. \textbf{ATSS} reveals the essential difference between anchor-free and anchor-based methods, that is, the selection of positive and negative samples, then it proposed a more reasonable sample selection strategy. FCOS and ATSS have implemented Anchor-Free by a similar approach, making them face similar problems. As shown in Figure \ref{Detect_result}, we show as many anchor boxes as possible to ensure that the experimental conclusions are complete. It can be seen that FCOS and ATSS do not perform well in regression anchor boxes, both of which failed to enclose the metal pistol in all of their anchor boxes, and many anchor boxes have low confidence. Meanwhile, in the ATSS detection results, there is a confusion between the gaps and the object resulting in false alarms as we mentioned in Section \ref{Sec_dataset_and_detail}.

\textbf{PVT} is a famous Transformer-based backbone, which combines the advantages of CNNs and Transformers. It introduces a step-by-step contraction pyramid to obtain multi-scale output similar to CNN. Meanwhile, thanks to the self-attention mechanism in Transformer, PVT maintains the global receptive field at different scales. \textbf{Dyhead} is a novel detection head with the attention mechanism. It combines multiple self-attention mechanisms between feature levels, spatial locations, and output channels, and thus significantly improves the representation ability of object detection heads without the heavy computational overhead. Introducing the attention mechanism makes the results of PVT and Dyhead very competitive, being only 7.3\% and 7.1\% lower than ours in mAP, respectively. However, as shown in Figure \ref{Detect_result}, since the attention mechanism of PVT and Dyhead was carried out on a single head, they had some wrong classification in the detection results. And the performance comparison of all these models is visualized in Figure \ref{points_SOTA}.

\subsection{Ablation Study}
\label{sec_ablation}

In this section, we conduct a series of ablation studies to evaluate the contribution of the proposed modules. In the Tables \ref{ablation}$\sim$\ref{ablation_QSE}, we use the terms "QSE", "QSH", "DAM", "DAB", and "DN" to denote "Query Selection", "Query Sharing", "Deformable Attention Module", "Dynamic Anchor Boxes", and "DeNoising learning", and the "R50", "Swin-T", and "DCFT" to denote "Resnet-50", "Swin Transformer-Tiny", "Denoising Coarse-to-Fine Transformer", respectively.

\begin{table*}[!ht]
\centering
\begin{threeparttable}
\caption{\bf \textcolor{black}{Ablation study of the proposed algorithm components. }}
\label{ablation}
\renewcommand{\arraystretch}{1.2}
\setlength\tabcolsep{2mm}{
\begin{tabular}{lccccccccc}
\hline
\hline
\#Row              & QSH & QSE & Backbone & AP$_{50}$ & AP$_{75}$ & AP  & AP$_{S}$  & AP$_{M}$   & AR$_{100}$ \\
\hline
1. DETR(baseline)           &           &           & R50      & 92.9 & 63.6 & 49.1 & 45.7 & 54.7 & 56.1  \\
2. Strong baseline\tnote{1} &           &           & R50      & 95.8 & 72.4 & 58.0 & 56.9 & 63.8 & 63.1      \\
3. PMMW-DETR                &\checkmark &           & R50      & 97.0 & 86.8 & 65.7 & 42.5 & 67.4 & 71.9  \\
4. PMMW-DETR                &\checkmark &\checkmark & R50      & 97.1 & 89.3 & 68.7 & 44.4 & 70.5 & 77.1  \\
5. PMMW-DETR(Ours)          &\checkmark &\checkmark & DCFT     & \textbf{97.5}     & \textbf{91.8}     & \textbf{71.3}      & \textbf{49.1}      & \textbf{74.1}      & \textbf{80.4}   \\
\hline
\hline
\end{tabular}
}
\begin{tablenotes}   
    \footnotesize              
    \item[1] The strong baseline is built on DETR with modules of DAM, DAB, and DN.
\end{tablenotes} 
\end{threeparttable}     
\end{table*}

\begin{figure}[htp]
\centering
\includegraphics[width=8.6cm]{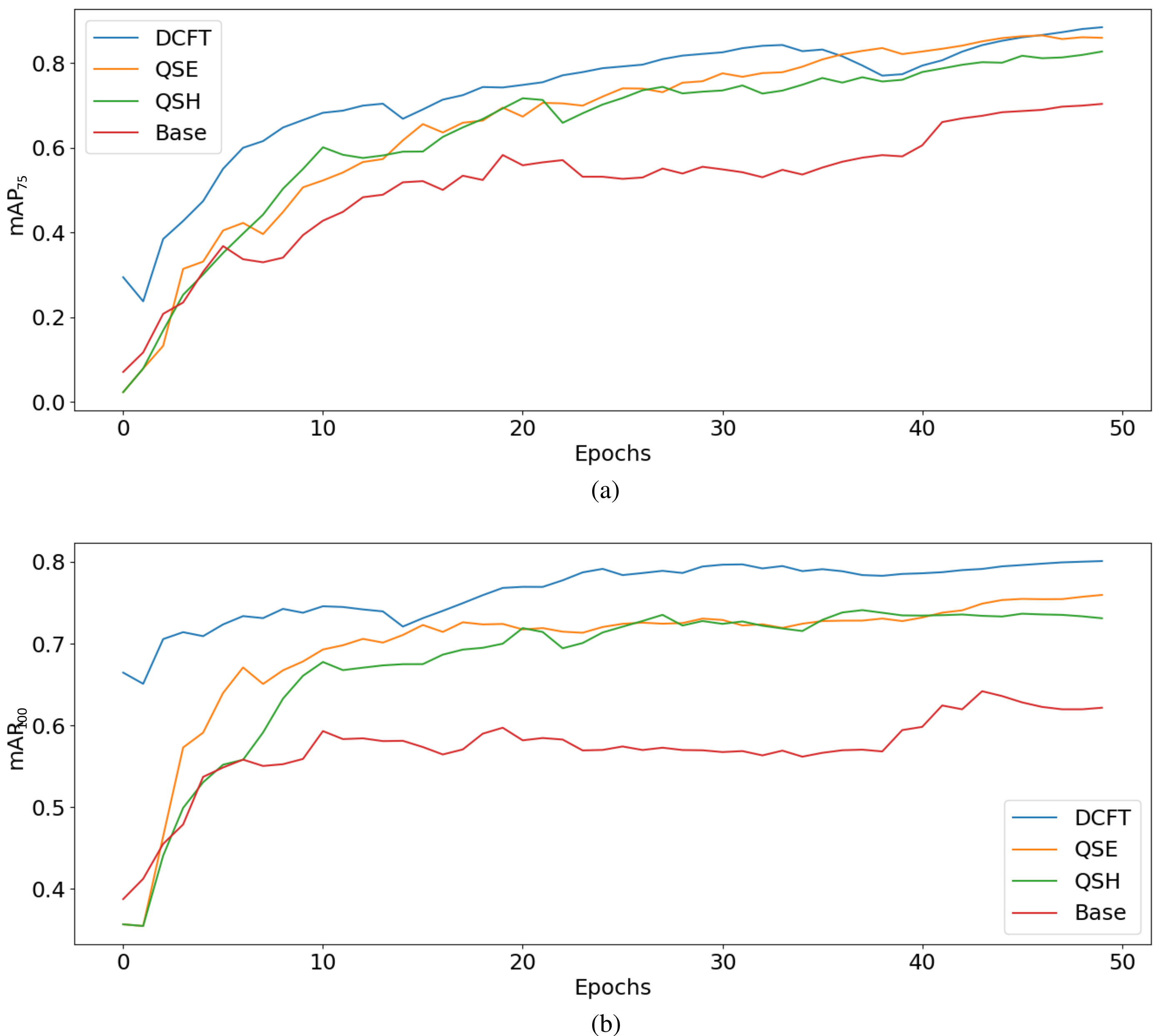}
\caption{The mAP$_{75}$ (a) and mAR$_{100}$ (b) curves for each component of PMMW-DETR.}
\label{curve_ablation}
\end{figure}

We use the DETR as the baseline, and build a strong baseline with modules of DAM, DAB, and DN. Then we add QSE, QSH, and DCFT to the strong baseline in proper order to verify the performance of our proposed modules. The results are available in Table \ref{ablation} Rows 3$\sim$5. In addition, Figure \ref{curve_ablation} shows the performance curves of each component. We can see that QSE, QSH, and DCFT modules improve performance significantly.

\begin{table}[!ht]
\centering
\caption{\bf \textcolor{black}{Comparison between different query initialization methods. }}
\label{ablation_QSE}
\renewcommand{\arraystretch}{1.2}
\setlength\tabcolsep{1.5mm}{
\begin{tabular}{lcccccc}
\hline
\hline
Method        & AP$_{50}$ & AP$_{75}$ & AP  & AP$_{S}$  & AP$_{M}$   & AR$_{100}$ \\
\hline
Static        & 97.0 & 86.8 & 65.7 & 42.5 & 67.4 & 71.9  \\
Dynamic       & 97.1 & 88.3 & 67.9 & 43.1 & 68.9 & 75.8  \\
QSE           & \textbf{97.1} & \textbf{89.3} & \textbf{68.7} & \textbf{44.4} & \textbf{70.5} & \textbf{77.1} \\
\hline
\hline
\end{tabular}
}
\end{table}

To verify the effectiveness of query selection and the impact of dynamic initialization, using static initialization as a baseline, we compare the performance of dynamic initialization and query selection. As can be seen from the Table \ref{ablation_QSE}, we can find our query selection method outperforms other methods. The dynamic initialization method indeed limits the performance of the network to a certain extent.

\begin{table}[htp]
\centering
\caption{\bf \textcolor{black}{Comparison between different backbones. }}
\label{ablation_backbone}
\renewcommand{\arraystretch}{1.2}
\setlength\tabcolsep{1mm}{
\begin{tabular}{lcccccccc}
\hline
\hline
backbone        & \#Params(M)   & AP$_{50}$ & AP$_{75}$ & AP  & AP$_{S}$  & AP$_{M}$   & AR$_{100}$ & FPS \\
\hline
R50          & 37.7  & 95.8 & 72.4 & 58.0 & 56.9 & 63.8 & 63.1 & 30 \\
PVT          & 34.2  & 96.1 & 80.4 & 64.0 & 39.8 & 65.3 & 69.1 & 26 \\
Swin-T       & 38.5  & 97.4 & 87.1 & 68.7 & 43.9 & 70.4 & 75.5 & 20 \\
DCFT         & 39.4  & \textbf{97.5}     & \textbf{91.8}     & \textbf{71.3}      & \textbf{49.1}      & \textbf{74.1}      & \textbf{80.4} & 24  \\
\hline
\hline
\end{tabular}
}
\end{table}

On the basis of Table \ref{ablation} Row 7, we replaced the DCFT with different backbones and compared their performance, and the results are shown in Table \ref{ablation_backbone}. The parameters and  FPS are calculated using the same network (RetinaNet). Since the multi-scale calculation is introduced in the backbone, the parameters of our proposed DCFT are slightly higher. However, our model outperforms mAP and mAR, which is important for security. And as we mentioned in Section \ref{Sec_backbone} the computational complexity of our model is lower than Swin Transformer, which leads to our FPS being higher than Swin-T. Due to the excessive computation of full map attention, it can only perform the image classification task that is less computational overhead but cannot complete the object detection task. Therefore, we added the PVT in the SOTA model to compare. The AP of R50 is poor because of the lack of global information. Similar to R50, PVT loses more long- and short-range information in the cascading reduction process, which results in the lower AP of PVT. Swin-T achieved competitive results with its compact design. However, its diminished distance to information resulted in 4.7\% lower in mAP$_{75}$ and 4.9\% lower in mAR$_{100}$ than ours. 

\subsection{hyperparameters setting for SOTA models}

In order to explore the relationship between the training hyperparameter settings of PMMW concealed object detection and optical image detection, we organized some comparative experiments represented by ATSS in this section.

\begin{table}[htp]
\centering
\caption{\bf Comparison between different hyperparameters settings. }
\label{hyperparameters_settings}
\setlength{\tabcolsep}{6pt} 
\renewcommand{\arraystretch}{1.2}
\setlength\tabcolsep{1mm}{
\begin{tabular}{lcccccccc}
\hline
\hline
Learning rate &\multicolumn{2}{c}{10$^{-1}$} &\multicolumn{3}{c}{10$^{-2}$} &\multicolumn{3}{c}{10$^{-3}$}\\
\hline
Momentum & 0.9 & 0.5 & 0.99 & 0.9 & 0.5 & 0.99 & 0.9 & 0.5  \\
\hline
AP$_{75}$ & NaN & 55.4 & NaN & \textbf{74.1} & 47.6 & 55.1 & 64.4 & 57.3 \\ 
\hline
\hline
\end{tabular}
}
\end{table}

We initially follow the setting of the original paper and mmdetection. We first tested the combined influence of Learning rate and Momentum on AP$_{75}$, the initial setting of Learning rate and Momentum is 10$^{-2}$ and 0.9. Momentum is set to 0.5, 0.9, and 0.99, which represent 2 times, 10 times, and 100 times the maximum speed of the original gradient descent, respectively. We take the example data from ATSS, the other models tend to be similar. And as shown in Table \ref{hyperparameters_settings}, we found that most of the original settings were consistent with the optimal results of the grid search. When we increase the learning rate to 10$^{-1}$ without reducing momentum, the model will experience gradient vanishing or gradient exploding during the training process, which is NaN in the table. When we then drop the Momentum to 0.5, the model finished 2x scheduled training but performance drops dramatically. We also tried reducing the learning rate, we think that the lower learning rate will lead to insufficient convergence. Changing the momentum will exacerbate the insufficiency, causing the AP$_{75}$ to be much lower than the original setting.

For the coefficients setting of the loss function, like Focal loss, we also explore the impact of hyperparameters. When we adjust the coefficient $\gamma$ in the range of $2 \times 10^{-2}$, the AP$_{75}$ changes only slightly. And we find the performance dramatically dropped when we vary the coefficients of the original paper beyond $5 \times 10^{-2}$. 

Since the training strategy affects different neural networks similarly, the other models also exhibit similar distributions despite differences in hyperparameter settings between them. At the same time, due to the time-consuming training of deep learning and the complexity of grid search, we often get relatively optimal solutions.

\section{Conclusion}

In this paper, we have presented a robust task-aligned Detection Transformer named PMMW-DETR for PMMW images with low resolution, sparse texture, and high noise. Different from popular neural networks, we perform the denoising coarse-to-fine Transformer backbone to deal with the problem of high noise and lack of texture information in the PMMW image, the query selection module to introduce spatial prior information, and the task-aligned dual-head block to enhance the ability of the network to identify object categories. Numerical experiments show that PMMW-DETR outperforms other SOTA methods. As far as we know, our network achieves the classification of different concealed objects for the first time. Moreover, PMMW-DETR is easy to generalize to other degraded image data where objects have prior location information. For the deployment and generalization ability of concealed object detection in the future, we will explore the potential of incremental learning and few-shot learning. Few-shot learning aims to maintain the same performance with few training samples, by pre-trained models or feature distillation, etc. Incremental learning focus on updating parameters based on the samples out of the distribution of the training set so that the model could renew knowledge in the working environment.

\bibliographystyle{IEEEtran}

\bibliography{sample}

\begin{IEEEbiography}[{\includegraphics[width=1in,height=1.25in,clip,keepaspectratio]{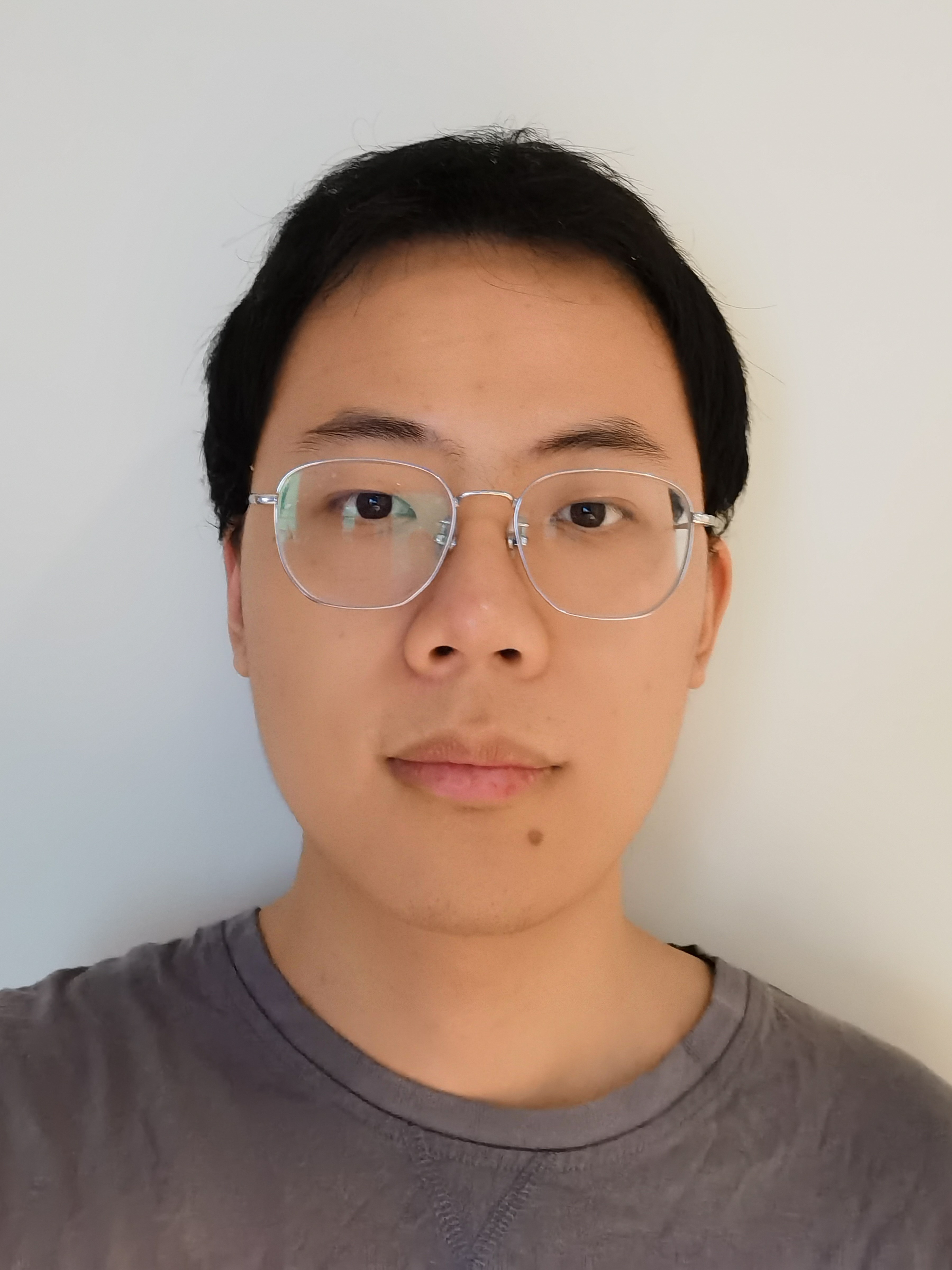}}]{Cheng Guo} was born in Hubei, China. He is currently pursuing the M.S. degree at the Huazhong University of Science and Technology (HUST), Wuhan, China. He received the B.E. degree in Electronic Information engineering from Wuhan University of Science and Technology, Wuhan, China, in 2021. 

His research interests include image processing, passive imaging, and object detection.
\end{IEEEbiography}
 
\begin{IEEEbiography}[{\includegraphics[width=1in,height=1.25in,clip,keepaspectratio]{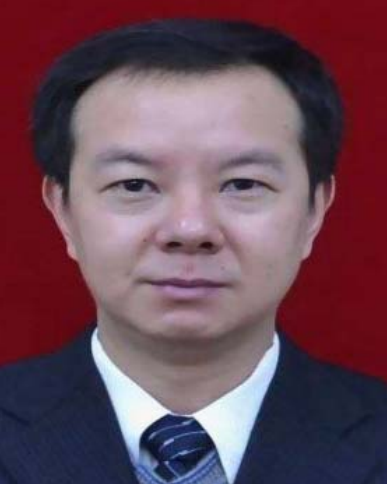}}]{Fei Hu} was born in Hubei, China. He received the M.S. degree in experimental mechanics and the Ph.D. degree in communication and information system from the Huazhong University of Science and Technology (HUST), Wuhan, China, in 1998 and 2002, respectively. 

He was a Senior Researcher with the National Key Laboratory of Science and Technology on Multi-Spectral Information Processing Technologies, HUST. He is currently a Professor with the School of Electronic Information and Communications, HUST. He has authored or coauthored more than 40 technical articles. His research interests include microwave technique, microwave interferometric radiometers, microwave remote sensing, and passive microwave imaging.
\end{IEEEbiography}

\begin{IEEEbiography}[{\includegraphics[width=1in,height=1.25in,clip,keepaspectratio]{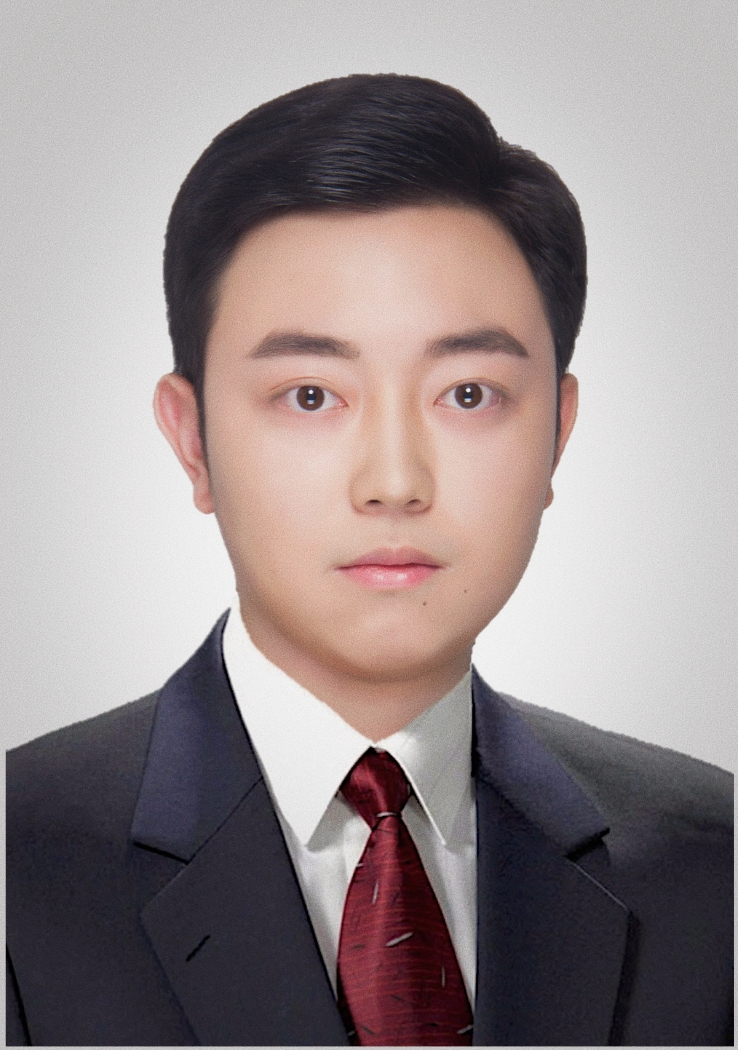}}]{Yan Hu} was born in Hubei, China in 1995. He received the B.S. degree in electronic and information engineering from the Huazhong University of Science and Technology (HUST), Wuhan, China, in 2017. He received his Ph.D. degree from HUST, in 2022. 

His research interests include passive imaging, polarization characteristics, and image processing.
\end{IEEEbiography}

\end{document}